\documentclass[letterpaper, 10 pt, conference]{ieeeconf}  

\usepackage{amsmath} 
\usepackage{amssymb}  
\usepackage{caption}
\usepackage{algorithm}
\usepackage{algpseudocode}
\usepackage{booktabs}
\usepackage{multirow}
\usepackage[bookmarks=true]{hyperref}
\usepackage{xcolor}
\usepackage{colortbl}
\let\labelindent\relax
\usepackage{enumitem}
\usepackage{amsmath}
\DeclareMathOperator*{\argmin}{arg\,min}
\usepackage{marvosym}
\usepackage[table]{xcolor}
\definecolor{AcadGreen}{HTML}{2E7D32} 
\definecolor{AcadRed}{HTML}{C62828}   
\definecolor{AcadBlue}{HTML}{1565C0} 
\newcommand{\deltag}[1]{\textcolor{AcadGreen}{(#1\%)}}
\newcommand{\deltar}[1]{\textcolor{AcadRed}{(#1\%)}}

\algrenewcommand\algorithmiccomment[1]{\hfill \textcolor{gray}{\textit{#1}}} 

\usepackage[colorinlistoftodos,prependcaption,textsize=tiny]{todonotes}

\title{\LARGE \bf
RoboOS-NeXT: A Unified Memory-based Framework for Lifelong, Scalable, and Robust Multi-Robot Collaboration
}

\author{\small \textbf{Huajie Tan$^{1,2,*}$, Cheng Chi$^{2,*}$, Xiansheng Chen$^{2,*}$, Yuheng Ji$^{2,3,*}$, Zhongxia Zhao$^{2}$, Xiaoshuai Hao$^{2}$,} \\
\small  \textbf{Yaoxu Lyu$^{1,2}$, Mingyu Cao$^{2}$, Junkai Zhao$^{2}$, Huaihai Lyu$^{2,3}$, Enshen Zhou$^{2,4}$, Ning Chen$^{1,2}$, Yankai Fu$^{1,2}$,} \\
 \textbf{Cheng Peng$^{2,3}$, Wei Guo$^{2}$, Dong Liang$^{2}$, Zhuo Chen$^{2}$, Mengsi Lyu$^{2}$, Chenrui He$^{2}$, Yulong Ao$^{2}$,} \\
 \textbf{Yonghua Lin$^{2}$, Pengwei Wang$^{2,\dagger}$, Zhongyuan Wang$^2$, Shanghang Zhang$^{1,2,\text{\Letter}}$} \\
$^1$ \small State Key Laboratory of Multimedia Information Processing, School of Computer Science, Peking University \\ 
$^2$ \small Beijing Academy of Artificial Intelligence
$^3$ \small Institute of Automation, Chinese Academy of Sciences 
$^4$ \small Beihang University
}

\begin{document}

\twocolumn[{%
\renewcommand\twocolumn[1][]{#1}%
\maketitle
\begin{center}
    \centering
    \captionsetup{type=figure}
    \includegraphics[width=0.98\linewidth]{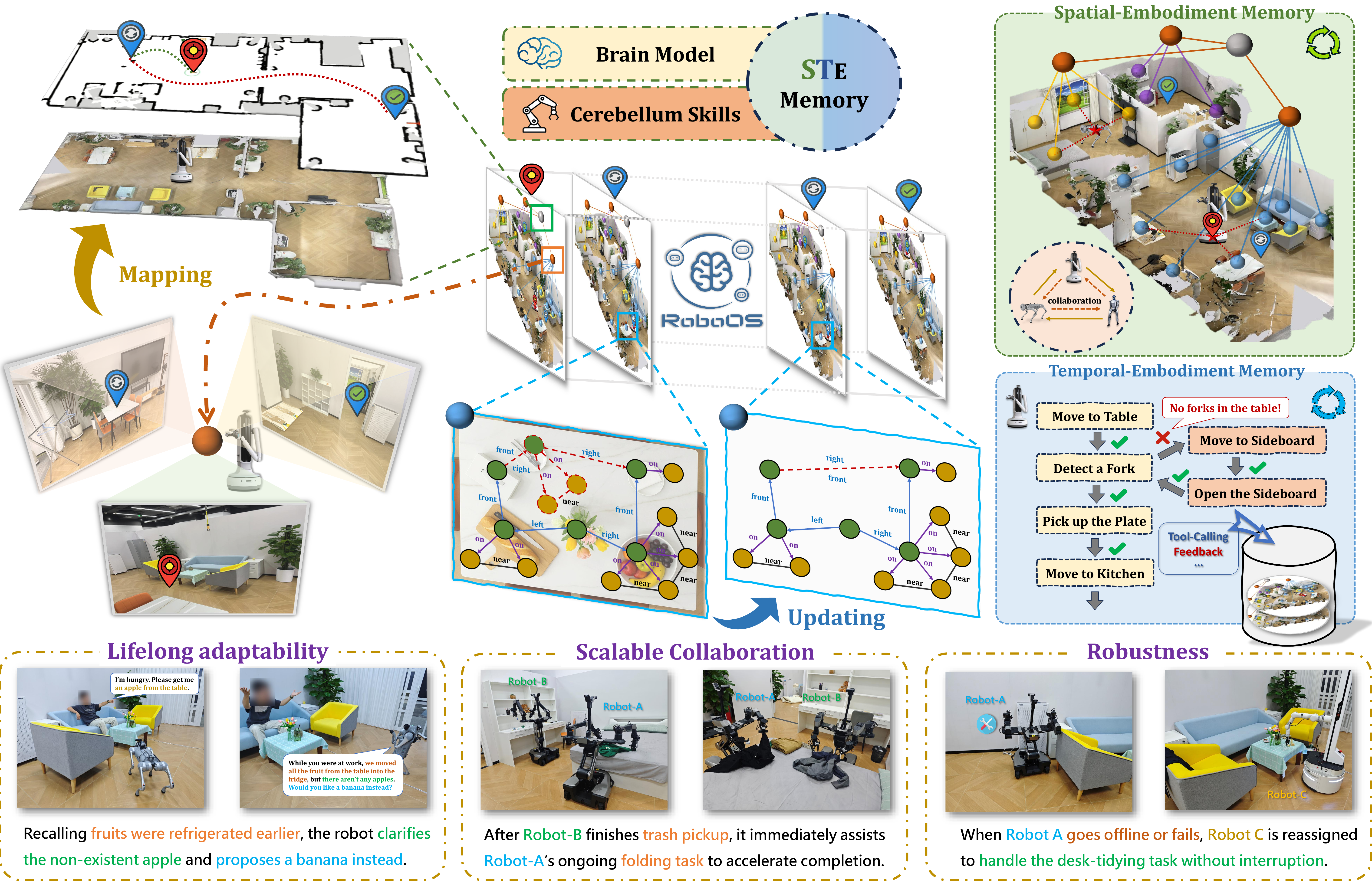}
    \captionof{figure}{
    \textbf{Overview of RoboOS-NeXT.} RoboOS-NeXT is a unified memory-based framework for multi-robot collaboration, built around a shared Spatio-Temporal–Embodiment Memory (STEM). STEM provides a unified representation by integrating spatial scene geometry, temporal event history, and embodiment profiles, making it accessible to all robots. Based on the STEM, a brain–cerebellum framework closes the loop between cognition, planning and control, supporting lifelong adaptation, scalable collaboration and robust scheduling.
}

    \label{fig:intro}
\end{center}
}]

\begin{abstract}
The proliferation of collaborative robots across diverse tasks and embodiments presents a central challenge: achieving lifelong adaptability, scalable coordination, and robust scheduling in multi-agent systems. Existing approaches, from vision-language-action (VLA) models to hierarchical frameworks, fall short due to their reliance on limited or dividual-agent memory. This fundamentally constrains their ability to learn over long horizons, scale to heterogeneous teams, or recover from failures, highlighting the need for a unified memory representation. To address these limitations, we introduce \textbf{RoboOS-NeXT}, a unified memory-based framework for lifelong, scalable, and robust multi-robot collaboration. At the core of RoboOS-NeXT is the novel \textbf{Spatio-Temporal–Embodiment Memory (STEM)}, which integrates spatial scene geometry, temporal event history, and embodiment profiles into a shared representation. This memory-centric design is integrated into a brain-cerebellum framework, where a high-level brain model performs global planning by retrieving and updating STEM, while low-level controllers execute actions locally. This closed loop between cognition, memory, and execution enables dynamic task allocation, fault-tolerant collaboration, and consistent state synchronization. We conduct extensive experiments spanning complex coordination tasks in restaurants, supermarkets, and households. Our results demonstrate that RoboOS-NeXT achieves superior performance across heterogeneous embodiments, validating its effectiveness in enabling lifelong, scalable, and robust multi-robot collaboration. Project website: \href{https://flagopen.github.io/RoboOS/}{RoboOS-NeXT}.

\end{abstract}
\section{INTRODUCTION}
The vision of a home maintained by autonomous robots, which patrol, detect clutter, and collaboratively restore order, illustrates the promise of embodied intelligence in everyday environments.
This vision hinges on three fundamental properties of embodied systems: 
\textit{\textbf{lifelong adaptability}} for continual accumulation and reuse of prior experience; 
\textit{\textbf{scalable collaboration}} for orchestrating collaboration across large and diverse robot collectives; 
and \textit{\textbf{robustness}} for maintaining stability in dynamic or failure-prone environments~\cite{amazon_robotics_2025, mandi2024roco, an2023multi, liu2024coherent,tan2025roboos}.
Achieving these properties requires systems that can proactively maintain order by leveraging past experience, dynamically orchestrate multiple agents for complex tasks, and reliably recover from unexpected challenges such as hardware malfunctions or ambiguous user commands.
These three aspects are exemplified by the scenarios of lifelong adaptation, scalable collaboration, and robust scheduling, as illustrated in Fig.~\ref{fig:intro}.

Despite recent progress, current approaches remain insufficient to realize this vision. End-to-end vision-language-action (VLA) models advance robot learning by directly mapping perception to action~\cite{kim2024openvla, liu2024rdt, black2024pi_0, Helix2024, cui2025openhelix, team2025gemini_robo, bjorck2025gr00t,liu2024robomamba2, li2025language,li2024lamp,li2025omnimotion}, but they rely on scarce training data and exhibit low sample efficiency, limiting generalization across embodiments, environments, and tasks. 
Hierarchical frameworks improve controllability through task decomposition and modular reasoning~\cite{vemprala2024chatgpt, shi2025hi, pi05, huang2023voxposer}, yet they remain individual-agent centric and scale poorly to multi-robot settings, their policies are tightly coupled to specific morphologies and thus fragile under embodiment changes, and they lack persistent memory to support lifelong adaptation.

{\textit{These limitations highlight the need for embodied systems equipped with memory.}} While recent studies explore memory via 3D scene graphs~\cite{hu2024hiagent, gu2024conceptgraphs}, cached states for long-horizon tracking~\cite{fan2024embodied}, or structured grounding and program synthesis~\cite{ahn2022can, liang2022code}, such approaches provide only incremental improvements, often confined to single robots or short-lived contexts. 
\textit{What is still missing is a unified representation that integrates spatial, temporal, and embodiment memory to enable lifelong, scalable, and robust multi-robot collaboration.}

To address these challenges, we propose \textbf{RoboOS-NeXT}, a unified memory-based framework for multi-robot collaboration, built on the \textit{Spatio-Temporal–Embodiment Memory (STEM)}.
STEM provides a unified representation of spatial, temporal, and embodiment dimensions, and the interactions within this representation enable lifelong adaptation, scalable collaboration, and robust scheduling:
\textit{\textbf{(1) Spatial.}} STEM encodes multi-view 3D geometry that represents the global scene structure, and dynamic scene graphs that model object–object and object–robot relations.
\textit{\textbf{(2) Temporal.}} It tracks the evolution of system states, including object transitions, task progress with feedback, and operational logs, thereby maintaining execution context.  
\textit{\textbf{(3) Embodiment.}} It profiles heterogeneous robots across their lifecycle, encompassing accumulated experience, current perceptual–execution states, and available resources.  
This unified representation enables cross-dimensional interactions: spatio–temporal integration models evolving environments, temporal–embodiment integration facilitates experience sharing across robots, and spatio–embodiment integration ensures consistency in collaboration. Together, these mechanisms establish a \emph{continuous, extensible, reliable} memory foundation for lifelong adaptation, scalable collaboration, and robust scheduling.

On this basis, RoboOS-NeXT integrates STEM with a \emph{brain–cerebellum} hierarchical framework to link global reasoning and local execution. 
The brain invokes and updates STEM for high-level reasoning and task decomposition, while the cerebellum performs low-latency actions and local corrections guided by memory.    
This closed loop of cognition, execution, and memory synchronizes states across robots, enables dynamic task allocation, and supports fault-tolerant collaboration, thereby realizing lifelong adaptation, scalable collaboration, and robust scheduling.  
The contributions of this paper are summarized as follows:  

\begin{itemize}  
\item We present \textbf{RoboOS-NeXT}, a memory-based framework for multi-robot collaboration, built on STEM, which integrates spatial, temporal, and embodiment dimensions into a unified representation;

\item We design a \textbf{Brain–Cerebellum–Memory} hierarchical loop that connects global reasoning with skill execution through STEM, providing a principled basis for multi-robot collaboration;  

\item We evaluate RoboOS-NeXT on diverse tasks in restaurants, households, and supermarkets, complemented by real-world demonstrations, demonstrating its effectiveness across heterogeneous embodiments.
\end{itemize}  
\section{RELATED WORK}

\subsection{Embodied Vision–Language Models}
Recent advances in vision–language models (VLMs) have greatly improved perception, grounding, and reasoning across visual and textual modalities~\cite{zhou2025roborefer,han2025tiger,luo2025robobench}. Closed-source systems such as GPT-4o~\cite{hurst2024gpt4o}, Claude-3.5~\cite{anthropic2024claude3.5}, and Gemini~\cite{Google2023gemini}, along with open-source counterparts~\cite{Qwen2.5-VL, chen2024internvl, li2024llavaov, an2025llava}, have achieved strong performance in VQA, captioning, and dialogue understanding. Reasoning-enhanced variants such as GPT-o1~\cite{OpenAI2024o1}, DeepSeek-R1~\cite{guo2025deepseek}, and Kimi-1.5~\cite{team2025kimi}, as well as reinforcement-tuned models~\cite{tan2025reason,huang2025vision,li2025language}, further extend multi-step reasoning and cognitive consistency. Building on these developments, embodied VLMs have emerged to integrate such multimodal reasoning into robotics, treating them as “embodied brains.” Early systems such as EmbodiedGPT~\cite{mu2023embodiedgpt} and RoboBrain~\cite{ji2025robobrain} connect language-driven reasoning with robotic perception and control, while recent works including Robix~\cite{fang2025robix}, RynnEC~\cite{dang2025rynnec}, Ve-Brain~\cite{luo2025vebrain}, and RoboBrain-2.0~\cite{team2025robobrain} pursue unified architectures that couple perception, reasoning, and planning within a single model. Recent efforts have also begun emphasizing spatial intelligence, which enables embodied models to reason over 3D geometry, object relations, and scene dynamics for more grounded manipulation and navigation~\cite{bai2025towards,lyu2025egoprompt,ji2025visualtrans,li2024mlip,song2025maniplvm,ji2025enhancing, zhang2025vcot, li2024mulsmo, li2025manipdreamer3d,liu2024segment,zhang2025beyond}. Despite this progress, these embodied VLMs remain constrained by limited long-term memory, embodiment transferability, and real-time responsiveness, preventing them from achieving lifelong learning, scalable collaboration, and robust execution. In response, RoboOS-NeXT couples a unified memory system with a Brain–Cerebellum–Memory loop, tightening the link between reasoning and control.

\subsection{Architectural Paradigms for Embodied Control}

Research on embodied control has largely followed two architectural paradigms.  
The first is Vision–Language–Action (VLA) models, which map perceptual and linguistic inputs directly to robot actions.  
Progress in this direction has been driven by scaling real-robot demonstrations and coupling them with web-scale vision–language pretraining.  
Representative systems such as the RT series~\cite{brohan2023rt,zitkovich2023rt}, OpenVLA~\cite{kim2024openvla}, pi0~\cite{black2024pi_0}, Gemini Robotics~\cite{team2025gemini_robo}, and related efforts~\cite{liu2024rdt,liu2024robomamba,ji2025robobrain, bai2025rethinking, fan2025long,li2023enhancing, li2023general,wu2024robomind,yuan2025motiontrans} demonstrate the potential of this approach, moving toward more generalist policies.
Together, these advances position VLAs as a promising paradigm for embodied control, while still being heavily data-hungry, sample-inefficient for long-horizon or contact-rich tasks, and lacking persistent memory or shared context across tasks and agents.
The second paradigm, hierarchical frameworks, introduces task decomposition and modular reasoning to address some of these limitations.  
Representative examples include VoxPoser~\cite{huang2023voxposer}, which leverages compositional 3D value maps for manipulation, and recent systems that integrate large language models as high-level planners with low-level controllers~\cite{vemprala2024chatgpt,shi2025hi,pi05,zhou2025code,huang2025rekep}.  
These designs improve controllability and robustness by isolating subproblems, but they often lack persistent shared memory across tasks, limit coordination to individual agents, and show brittle performance under embodiment changes or long-horizon demands.  
Beyond task decomposition, recent frameworks have begun to incorporate memory to improve embodied control.  
Approaches such as retrieval-augmented agents, snapshot-based 3D scene memories, open-vocabulary scene graphs, and working-memory modules~\cite{zhu2024retrieval,yang2024snapmem,gu2024conceptgraphs,fan2024embodied} demonstrate the benefits of memory augmentation for spatial grounding, temporal consistency, and long-horizon reasoning.  
Yet these remain largely constrained to single-agent or episodic contexts, and what is still missing is a unified memory representation that enables lifelong adaptation, scalable collaboration, and robust scheduling.

\subsection{Multi-Robot Collaboration}

Multi-robot collaboration (MRC) has a long history in robotics, spanning domains such as automated warehousing~\cite{agrawal2023rtaw} and search and rescue~\cite{guo2023cross}.  
Classical approaches focused on coordination protocols, task allocation, and communication strategies~\cite{rizk2019cooperative,fierro2018multi}, typically assuming homogeneous teams and structured environments.  
Learning-based methods, including multi-agent reinforcement and imitation learning~\cite{patino2023learning,liu2023guide}, improved adaptability under uncertainty but continue to struggle with embodiment heterogeneity, dynamic re-planning, and real-time fault tolerance.  
More recent efforts have sought to bridge these gaps through shared memory for cooperative planning, fault-tolerant coordination under sensing or actuation failures, and collaborative manipulation in dynamic environments~\cite{sagirova2025srmt,aina2025fault,tan2025roboos,adil2025multi}.  
These advances demonstrate the potential of MRC systems to move beyond static protocols and adapt to uncertainty, yet they remain highly task-specific, often confined to navigation or manipulation.  
They rarely integrate high-level semantic reasoning with low-level execution, nor do they offer persistent, shared memory across agents to support long-term adaptation and synchronization.  
Consequently, current embodied control and multi-robot collaboration frameworks remain fragmented and fall short of providing the unified memory representation needed for lifelong adaptability, scalable collaboration, and robust scheduling in open-world environments.
\newcommand{\STM}{\mathcal{M}}
\newcommand{\SM}{\mathcal{S}}
\newcommand{\TM}{\mathcal{T}}
\newcommand{\EM}{\mathcal{E}}
\newcommand{\Reduce}{\operatorname{Reduce}}

\begin{figure*}[htbp]
    \centering
    \includegraphics[width=0.98\linewidth]{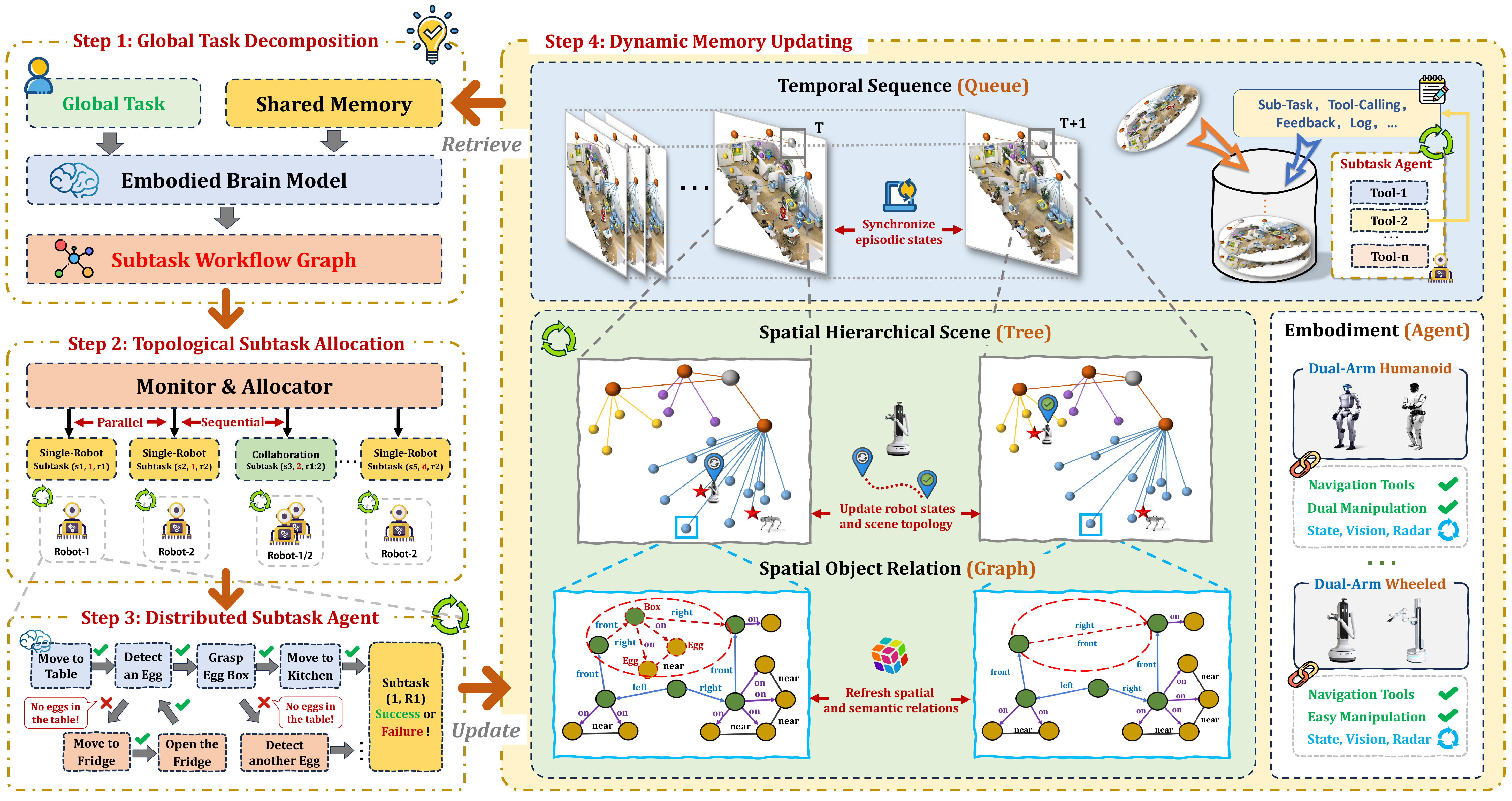}
    \caption{
        \textbf{Pipeline of RoboOS-NeXT.} The RoboOS-NeXT framework implements a workflow pipeline for multi-robot collaboration, consisting of four key phases: (1) global task decomposition, (2) topological subtask allocation, (3) distributed subtask agent, and (4) dynamic memory updating. Together, these phases establish a memory-centric workflow that enables lifelong, scalable, and robust multi-robot collaboration.
    }
    \label{fig:pipeline}
    \vspace{-1em}
\end{figure*}

\section{METHOD}
\subsection{Spatio-Temporal--Embodiment Memory (STEM)}
\label{sec:method:stem}

We introduce STEM as a unified memory representation that couples three complementary facets of task execution. At any time $t$, the memory state is defined as,
\begin{align}
\STM(t) = \big(\SM(t),\,\TM(t),\,\EM(t)\big),
\end{align}
where, $\STM$ is the full memory state; $\SM$ is \emph{Spatial Memory} (spatial geometry and semantics), $\TM$ is \emph{Temporal Memory} (event-level history with tool/feedback traces), and $\EM$ is \emph{Embodiment Memory} (robot capabilities, resources, and status).
The state evolves by a left-fold reduction over a time-stamped event stream:
\begin{align}
\STM(t) = \Reduce\!\big(\mathcal{U},\,\STM_0,\,\{e_k\}_{k=1}^{t}\big),
\end{align}
where $\Reduce$ applies the deterministic update operator $\mathcal{U}$ to initial state $\STM_0$ with a stream of events $\{e_k\}_{k=1}^{t}$.

Specifically, STEM is organized top--down as a \textit{queue--tree--graph--agent} structure: (1) the temporal \emph{queue} $\TM$ stores event records (\emph{when}); (2) the spatial \emph{tree-graph} $\SM$, including scene-level tree $\mathcal{S}_\mathrm{T}$ that captures root/region/carrier hierarchy (\emph{where}) and object-level graphs $\{\mathcal{S}_{G,c}\}$ that encode inter-object relations (\emph{what}); (3) the embodied \emph{agent} $\EM$ maintains robot nodes, their localization, capabilities, resources, sensors, and availability (\emph{who/how}).

\vspace{0.25em}
\noindent\textbf{Temporal Memory \textit{(Queue)}.}
We maintain an append-only, time-ordered list that logs state deltas, staged task context, and tool-call traces:
\begin{align}
\TM_i = \big[(\tau_i,\,\Delta\SM_i,\,\Delta\EM_i,\,g,\,\mathcal{Q}^{\text{pre}}_g,\,\mathcal{L}^{\text{tool}}_{\text{cur}})\big]_{i:\,\tau_i\le t},
\label{eq:temporal_log}
\end{align}
where $\tau_i$ denotes the event timestamp; $\Delta\SM_i$ is the spatial-memory variation at $\tau_i$ (e.g., object/relation insert, move, or delete); $\Delta\EM_i$ is the embodiment-memory variation at $\tau_i$ (e.g., capability/status/resource updates); $g$ is the global task identifier associated with this event; $\mathcal{Q}^{\text{pre}}_g$ is the pre-subtask queue for $g$ (pending subtasks that precede or enable the current subtask); and $\mathcal{L}^{\text{tool}}_{\text{cur}}$ is the tool-call log attached to the current subtask, which is expressed as follow:
\begin{equation}
\mathcal{L}^{\text{tool}}_{cur}
= \big[(\text{tool},\,\text{args},\,\text{status}\!\in\!\{\textsc{ok},\textsc{fail}\},\,\text{feedback})\big].
\end{equation}

\vspace{0.25em}
\noindent\textbf{Spatial Memory for Hierarchical Scene \textit{(Tree)}.}
We model the scene as a rooted, typed, multi-branch tree:
\begin{align}
\mathcal{S}_{\mathrm{T}}&=(\mathcal{V},\mathcal{E},r),\quad
\mathcal{V}= \mathcal{V}^{\mathrm{root}}\!\cup\!\mathcal{V}^{\mathrm{region}}\!\cup\!\mathcal{V}^{\mathrm{carrier}} .
\label{eq:scene_tree}
\end{align}
The root $r$ is the \emph{global scene}, maintaining a top-down 3D reconstruction and a 2D SLAM map in node $\mathcal{V}^{\mathrm{root}}$. Region nodes $\mathcal{V}^{\mathrm{region}}$ (e.g., each room in an apartment) store aligned multi-view imagery for specific region. Carrier nodes $\mathcal{V}^{\mathrm{carrier}}$ denote (im)movable supports (e.g., desk, dining table, planter). Each carrier anchors an object-level graph $\mathcal{S}_{G,c}$, which will be illustrated as follow.

\vspace{0.25em}
\noindent\textbf{Spatial Memory for Object Relation \textit{(Graph)}.}
Each carrier $c$ hosts $\mathcal{S}_{G,c}=(V_c,E_c)$, where each node $v \!\in\! V_c$ represents an object stored in the carrier, and each edge $e \!\in\! E_c$ encodes a spatial relation between two objects. Node $v\!\in\!V_c$ stores
\begin{align}
\mathbf{a}(v) = \big(\boldsymbol{\pi}_v,\;\boldsymbol{\sigma}_v,\;\mathbf{T}_v\big),
\label{eq:attr}
\end{align}
where $\boldsymbol{\pi}_v$ are intrinsic properties (category/size/affordances), $\boldsymbol{\sigma}_v$ dynamic states, and $\mathbf{T}_v$ the pose. Spatial relations use a typed predicate set,
\begin{align}
\mathcal{R}=\{\textsc{on},\textsc{in},\textsc{left},\textsc{right},\textsc{front},\textsc{back},\textsc{near}\},
\end{align}
with geometric predicates $\Phi_r$:
\begin{align}
E_c &\subseteq V_c \times \mathcal{R} \times V_c, \\
(v_1, rel, v_2) \!\in\! E_c 
&\iff \Phi_{rel}\big(\mathbf{T}_{v_1},\mathbf{T}_{v_2}\big)=\textsc{true}.
\end{align}
In each carrier’s local frame, we model objects as nodes with attributes/state/pose and connect them via approximate geometric relations, updating the graph with filtered observations for efficient querying and planning.

\vspace{0.25em}
\noindent\textbf{Embodiment Memory \textit{(Agent)}.} For each robot $r\!\in\!\EM(t)$ in the scene, we keep a profile
\begin{align}
\phi_r(t)=\big(\text{loc}_r(t),\,\mathcal{C}_r,\,\boldsymbol{\rho}_r(t),\,\mathbf{s}_r(t),\,\alpha_r(t)\big),
\label{eq:embodiment_profile}
\end{align}
where $\text{loc}_r$ links into the scene tree (region/carrier), $\mathcal{C}_r$ lists skills/tools (navigation, manipulation, special actions), $\boldsymbol{\rho}_r$ denotes resources (battery/CPU/net), $\mathbf{s}_r$ sensor snapshots (vision/tactile), and $\alpha_r\!\in\!\{\textsc{idle},\textsc{busy},\textsc{offline}\}$ indicates availability. Profiles are \emph{heartbeat-updated}: every $\Delta_H$ the robot emits a status event to refresh $\phi_r$. Tools are plug-and-play; capability changes produce typed update events.

\subsection{STEM Generation and Lifelong Update}
\label{sec:method:gen}

\noindent
\textbf{(1) Spatial Memory.}
\textcolor{AcadBlue}{\textit{Initialization.}} Given a new scene, we:
(i) reconstruct a global 3D point cloud and obtain a top-down view;
(ii) perform semantic segmentation/grounding on the point cloud to obtain carrier/object 3D boxes $\{\mathcal{B}_k\}$;
(iii) instantiate the scene tree $\mathcal{S}_{\mathrm{T}}$ by placing region and carrier nodes at $\mathrm{center}(\mathcal{B}_k)$ (task areas like rooms become region children of the root);
(iv) for each carrier node $c$, run multi-view scanning to detect/localize objects and populate its object-level scene graph $\mathcal{S}_{G,c}$;
(v) perform $\mathcal{V}^{\mathrm{root}}$--$\mathcal{V}^{\mathrm{region}}$ alignment by estimating the rigid transform from reconstruction to SLAM (Eq.~\eqref{eq:icp}) and registering each region’s multi-view to 3D via PnP (Eq.~\eqref{eq:pnp}):

\begingroup
\makeatletter
\def\tagform@#1{\maketag@@@{\normalsize(#1)}}
\makeatother
\small
\begin{equation}
T_{M\leftarrow P}^\star
= \argmin_{T\in SE(3)} \sum_{j}\big\|
\Pi_{\mathcal{M}}\!\big(T\mathbf{X}_j\big)-\mathbf{y}_j\big\|_2^2 ,
\label{eq:icp}
\end{equation}
\begin{equation}
(R_k,\mathbf{t}_k)^\star
= \argmin_{R\in SO(3),\,\mathbf{t}\in\mathbb{R}^3}
\sum_{j}\big\|\mathbf{u}_{k,j}-\pi\!\big(K(R\mathbf{X}_j+\mathbf{t})\big)\big\|_2^2 ,
\label{eq:pnp}
\end{equation}
\endgroup
where $\mathcal{P}=\{\mathbf{X}_j\in\mathbb{R}^3\}$ are 3D points from the reconstruction, 
$\mathcal{M}$ is the SLAM map, 
$\Pi_{\mathcal{M}}$ projects a 3D point into the SLAM/map frame,
$\mathbf{y}_j$ are matched 2D map keypoints in $\mathcal{M}$, 
$T_{M\leftarrow P}\!\in\!SE(3)$ is the rigid transform from reconstruction to SLAM,
$SE(3)$/$SO(3)$ denote the rigid/rotation groups,
$I_k$ is the $k$-th image with intrinsics $K$,
$\mathbf{u}_{k,j}\!\in\!\mathbb{R}^2$ are 2D image keypoints,
$\pi(\cdot)$ denotes perspective division,
and $(R_k,\mathbf{t}_k)$ is the camera pose of $I_k$.
This yields a consistent mapping: image $\rightarrow$ 3D $\rightarrow$ SLAM, enabling semantic localization and cross-view reasoning.
\textcolor{AcadBlue}{\textit{Updates (standard primitives).}}
Spatial edits act on $\mathcal{S}_{\mathrm{T}}$ and $\{\mathcal{S}_{G,c}\}$ using \textsc{Add}/\textsc{Remove}/\textsc{Move} primitives; each edit triggers relation re-evaluation locally:
\begingroup
\makeatletter
\def\tagform@#1{\maketag@@@{\normalsize(#1)}}
\makeatother
\small
\begin{align}
\textsc{Add}(\mathcal{S}_{G,c},v,\mathbf{a})\!:\;&
V_c \leftarrow V_c\cup\{v\},\\
& E_c \leftarrow E_c\cup\{(v_i,r,v_j)\}_{\Phi_r},\\[2pt]
\textsc{Remove}(\mathcal{S}_{G,c},v)\!:\;&
V_c \leftarrow V_c\setminus\{v\},\\
& E_c \leftarrow E_c\setminus\!\big(\{(v,*)\}\cup\{(*,v)\}\big),\\[2pt]
\textsc{Move}(\mathcal{S}_{G,c},v,\Delta\mathbf{T})\!:\;&
\mathbf{T}_v \leftarrow \Delta\mathbf{T}\circ \mathbf{T}_v,\\
& E_c \leftarrow \text{re-evaluate by }\Phi_r .
\end{align}
\endgroup
where $\mathbf{a}$ initializes the attributes of $v$ in Eq.\ref{eq:attr};\,
$\Delta\mathbf{T}\!\in\!SE(3)$ is an incremental rigid transform;\,
$\circ$ denotes transform composition (left action);\,
the subscript ${\Phi_r}$ indicates edges are recomputed via the predicate $\Phi_r$;\,
and $*$ is a wildcard, so $\{(v,*)\}\cup\{(*,v)\}$ removes all edges incident to $v$.

\medskip
\noindent\textbf{(2) Temporal Memory.}
We start with an empty, append-only, time-ordered queue $\TM(0)=\,[\,]$.
Every spatial edit or embodiment change emits an event into $\TM$.
The queue evolves by append:
\begin{align}
\TM(t{+}1) = \TM(t)\ \Vert\ \TM_i,
\end{align}
where $\TM_i$ has been defined in Eq.~\ref{eq:temporal_log} for event information.

\medskip
\noindent\textbf{(3) Embodiment Memory.}
For each robot $r\in\EM$ in the scene, we register a profile $\phi_r(0)$.
Embodiment memory is heartbeat-updated: every $\Delta_H$, robot $r$ emits a status event to refresh $\phi_r(t)$ (Eq.~\ref{eq:embodiment_profile}); sensor snapshots may update region multi-views and the SLAM map, and tool hot-plugging updates $\mathcal{C}_r$. During execution, ${loc}_r(t)$ snaps to the nearest region/carrier node (topological proximity in $\mathcal{S}_{\mathrm{T}}$), biasing allocation to the nearest capable robot.

\subsection{Brain–Cerebellum–Memory Framework}

The proposed RoboOS-NeXT demonstrates high task concurrency and flexibility in multi-robot task allocation. To clarify the overall workflow pipeline of RoboOS-NeXT, we use a single global task for detailed elaboration, as shown in Fig. \ref{fig:pipeline}.

\textbf{Step 1: Global Task Decomposition} Upon receiving the global task instruction \( T_{\text{global}} \), RoboOS-NeXT initiates a Retrieval-Augmented Generation (RAG) process via brain model to query the shared spatial memory, extracting environment-relevant information \( M_s \). This is integrated with (i) state feedback \( M_t \) from prior task executions (stored in shared temporal memory), (ii) the robots’ status-and-tool profile \( M_r \) (stored in shared embodiment memory), (iii) global task instruction \( T_{\text{global}} \). Brain model processes these inputs to generate a structured reasoning trace \( \mathcal{R} \) and a workflow graph \( \mathcal{G} \), which can be formalized as: 
\begin{equation}
    (\mathcal{R}, \mathcal{G}) = \text{BrainModel}\big(M_s \oplus M_t \oplus  M_r \oplus T_{\text{global}}\big),
\end{equation}
where \( \oplus \) denotes the concatenation or fusion of multimodal inputs, and \( \mathcal{G} \) can be expressed as follow:
\begin{equation} \mathcal{G} = \{[\text{s}_i, \text{d}_i,  \text{R}_i]\}_{i=1}^{n}, \end{equation}
where $n$ is the number of subtasks in the workflow, $\text{s}_i$ denotes the text description of $i^{th}$ subtask, $\text{R}_i\!\subseteq\!\EM$ is the assigned agent from the robot team, and $\text{d}_i\!\in\!\{0,1,2,\dots\}$ is the depth index (triples sharing the same $\text{order}$ run in parallel, and batches are dispatched non-decreasingly).

\textbf{Step 2: Topological Subtask Allocation} 
The Monitor dynamically schedules and allocates subtasks in parallel based on the topological dependencies encoded in the directed acyclic graph \( \mathcal{G} \). 
Each subtask in \( \mathcal{G} \) is classified into two types: \textit{(1) Single-Robot Subtask \((s, d, r_{p})\)}, executed autonomously by robot \( r_p \) at topological depth \( d \); and \textit{(2) Collaboration Subtask \((s, d, r_{p:q})\)}, requiring coordinated execution among multiple robots \(\{r_p, \dots, r_q\}\) at depth \( d \). 
To enforce dependency constraints, the Monitor employs \textit{Parallel Allocation}—executing independent subtasks concurrently at the same depth (\textit{e.g.}, \((s_1,1, r_1)\) and \((s_2, 1, r_2)\) in Fig. \ref{fig:pipeline})—and \textit{\textbf{Sequential Allocation}}, where subtask \((s_k, d_k, r_k)\) is blocked until all prerequisites at depth \( d_{k-1} \) are fulfilled (\textit{e.g.}, $(s_3, 2, r_{1:2})$ allocated after $(s_1,1, r_1)$ and $(s_2, 1, r_2)$). 
In practice, the system supports concurrent management of workflow graphs \(\{ \mathcal{G}_1, \mathcal{G}_2, \dots, \mathcal{G}_m \}\) for multiple global tasks, ensuring real-time adaptability to dynamic robot states and evolving task dependencies.

\textbf{Step 3: Distributed Subtask Agent}  
For each subtask, RoboOS-NeXT deploys a dedicated \textit{\textbf{Robotic Agent}} to manage execution. The Agent autonomously orchestrates tool selection from the Cerebellum Skill Library based on: (1) feedback from prior executions, (2) tool-calling history from temporal memory, and (3) robot-centric relation information (\textit{i.e.}, nearby nodes) from spatial memory of the scene. This closed-loop tool-calling facilitates dynamic error recovery.  
For example (Fig. \ref{fig:pipeline}), when robot are allocated with subtask (\textit{``Search for some eggs and place on the kitchen table''}), the Agent sequentially invokes tools (\textit{e.g., ``detect an egg''}). If the search fails (\textit{e.g., no egg detected in the dinning table}), the Agent uses spatial memory to infer potential locations (\textit{e.g., ``the fridge''}) and selects the navigation tool to \textit{``move to fridge''}, showcasing adaptive recovery through iterative tool refinement.

\textbf{Step 4: Dynamic Memory Updating}
Temporal memory and spatial memory are updated incrementally as robots perceive and act during subtask proceeding. Please also refer to subsec.\ref{sec:method:gen} for more details.

\section{Experiments}
\label{sec:experiments}

We design a comprehensive set of experiments to answer the following key research questions:

\begin{itemize}
    \item \textbf{RQ1 on Lifelong Adaptability:} How does RoboOS-NeXT's performance scale when faced with long-horizon, sequential tasks?
    \item \textbf{RQ2 on Collaborative Scalability:} How effectively does RoboOS-NeXT coordinate across an increasing number and diversity of robot embodiments?
    \item \textbf{RQ3 on Scheduling Robustness:} How robust is RoboOS-NeXT when facing environmental uncertainties and system faults?
    \item \textbf{RQ4 on Ablation:} What are the individual contributions of RoboOS-NeXT's core architectural components?
    \item \textbf{RQ5 on Failure Analysis:} What are the system's primary failure modes, and where do they occur in the execution pipeline?
\end{itemize}

\begin{table*}[t]
\centering
\caption{Evaluation of lifelong adaptability across varying sequence lengths (SQ = 1, 3, 5) and difficulty levels (L1–L3). Results are reported using MSR and AEST. Values in parentheses indicate relative change compared with the baseline.}
\label{tab:lifelong_eval}
\resizebox{\textwidth}{!}{%
\begin{tabular}{lc|*{4}{c}|*{4}{c}|*{4}{c}}
\toprule
\multirow{3}{*}[-1.2ex]{\textbf{Difficulty}} & \multirow{3}{*}[-1.2ex]{\textbf{SQ}} &
\multicolumn{4}{c|}{\textbf{Restaurant}} &
\multicolumn{4}{c|}{\textbf{Supermarket}} &
\multicolumn{4}{c}{\textbf{Household}} \\
\cmidrule(lr){3-6} \cmidrule(lr){7-10} \cmidrule(lr){11-14}
& & \multicolumn{2}{c}{MSR(\%)$\uparrow$} & \multicolumn{2}{c}{AEST(\#)$\downarrow$}
  & \multicolumn{2}{|c}{MSR(\%)$\uparrow$} & \multicolumn{2}{c}{AEST(\#)$\downarrow$}
  & \multicolumn{2}{|c}{MSR(\%)$\uparrow$} & \multicolumn{2}{c}{AEST(\#)$\downarrow$} \\
\cmidrule(lr){3-4}\cmidrule(lr){5-6}
\cmidrule(lr){7-8}\cmidrule(lr){9-10}
\cmidrule(lr){11-12}\cmidrule(lr){13-14}
& & Baseline & RoboOS-NeXT & Baseline & RoboOS-NeXT & Baseline & RoboOS-NeXT & Baseline & RoboOS-NeXT & Baseline & RoboOS-NeXT & Baseline & RoboOS-NeXT \\
\midrule
\multirow{3}{*}{\textbf{L1} (Simple)}
  & 1 & 76.6 & 80.8 \deltag{+4.2} & 19.2 & 14.3 \deltag{-26} 
      & 66.7 & 76.7 \deltag{+10.0} & 15.2 & 11.0 \deltag{-28} 
      & 81.6 & 89.2 \deltag{+7.6} & 18.3 & 11.6 \deltag{-37} \\
  & 3 & 22.5 & 77.5 \deltag{+55.0} & 18.8 & 14.7 \deltag{-22} 
      & 27.5 & 75.0 \deltag{+47.5} & 14.8 & 10.7 \deltag{-28} 
      & 27.5 & 90.0 \deltag{+62.5} & 19.1 & 11.1 \deltag{-42} \\
  & 5 & 0.0 & 79.2 \deltag{+79.2} & 18.4 & 13.8 \deltag{-25} 
      & 0.0 & 75.0 \deltag{+75.0} & 14.6 & 11.3 \deltag{-23} 
      & 4.2 & 87.5 \deltag{+83.3} & 17.5 & 11.9 \deltag{-32} \\
\midrule
\multirow{3}{*}{\textbf{L2} (Medium)}
  & 1 & 17.5 & 73.3 \deltag{+55.8} & 33.9 & 17.6 \deltag{-48} 
      & 19.2 & 73.3 \deltag{+54.1} & 26.1 & 13.6 \deltag{-48} 
      & 0.0 & 81.7 \deltag{+81.7} & 41.4 & 16.3 \deltag{-61} \\
  & 3 & 7.5 & 72.5 \deltag{+65.0} & 32.2 & 18.0 \deltag{-44} 
      & 5.0 & 70.0 \deltag{+65.0} & 25.0 & 13.0 \deltag{-48} 
      & 0.0 & 75.0 \deltag{+75.0} & 42.9 & 15.9 \deltag{-63} \\
  & 5 & 0.0 & 75.0 \deltag{+75.0} & 34.6 & 18.3 \deltag{-47} 
      & 0.0 & 66.7 \deltag{+66.7} & 29.6 & 14.2 \deltag{-51} 
      & 0.0 & 79.2 \deltag{+79.2} & 39.6 & 15.5 \deltag{-61} \\
\midrule
\multirow{3}{*}{\textbf{L3} (Complex)}
  & 1 & 0.0 & 67.5 \deltag{+67.5} & 99.7 & 27.1 \deltag{-73} 
      & 0.0 & 69.2 \deltag{+69.2} & 71.1 & 20.9 \deltag{-71} 
      & 0.0 & 60.0 \deltag{+60.0} & 82.1 & 24.3 \deltag{-70} \\
  & 3 & 0.0 & 62.5 \deltag{+62.5} & 96.5 & 28.1 \deltag{-71} 
      & 0.0 & 65.0 \deltag{+65.0} & 74.0 & 20.1 \deltag{-73} 
      & 0.0 & 55.0 \deltag{+55.0} & 84.4 & 23.3 \deltag{-72} \\
  & 5 & 0.0 & 66.7 \deltag{+66.7} & 102.1 & 27.8 \deltag{-73} 
      & 0.0 & 63.5 \deltag{+63.5} & 68.6 & 20.4 \deltag{-70} 
      & 0.0 & 58.3 \deltag{+58.3} & 79.9 & 23.2 \deltag{-71} \\
\bottomrule
\end{tabular}
}
\vspace{-1em}
\end{table*}

\subsection{Experimental Details}

\textbf{Scenario Setup.}
To evaluate RoboOS-NeXT at scale, we conduct experiments in a mock setting that abstracts away physical uncertainties and focuses on system effectiveness. The evaluation covers three domains: restaurants, supermarkets, and households, with 200 tasks instantiated in each. This setup enables controlled large-scale assessment of RoboOS-NeXT’s memory support and coordination capabilities, while complementary real-robot demonstrations serve as qualitative case studies in embodied environments.

\textbf{Evaluation Metrics.}
To comprehensively evaluate RoboOS-NeXT, we report a set of complementary metrics that jointly reflect effectiveness, efficiency, and robustness across different experimental settings:

\begin{itemize}
    \item \textbf{Success Rate (SR, \%)$\uparrow$:} The proportion of tasks successfully completed within the step budget. This serves as the primary measure of overall effectiveness and is reported in scalability, robustness, and ablations.
    
    \item \textbf{Marginal Success Rate (MSR, \%)$\uparrow$:} The success rate measured on the \emph{final task} of each lifelong or curriculum sequence. Unlike SR, which averages across all tasks, MSR reflects the ability to maintain stable performance across extended horizons without resets, and is thus critical for evaluating lifelong adaptation.
    
    \item \textbf{Average Execution Steps per Task (AEST, \#)$\downarrow$:} The average number of steps required to complete a task. Lower values indicate higher execution efficiency, and reductions in AEST across sequence lengths serve as evidence of experience reuse and adaptive learning.
    
    \item \textbf{Success per Step (SS, \%/\#)$\uparrow$:} Defined as the ratio between task success rate and the average number of steps, SS reflects the \emph{average accuracy achieved per step}. It provides a normalized measure that captures how effectively each action contributes to overall success.
\end{itemize}

\begin{table*}[t]
\centering
\caption{Scalability evaluation across different team compositions (SQ=1). 
We report AEST (lower is better) and SR/SS (higher is better). 
Wheel. denotes wheeled robots, Hum. denotes humanoids, and Quad. denotes quadrupeds.}
\label{tab:scalability_eval}
\resizebox{0.95\textwidth}{!}{%
\begin{tabular}{l|ccc|cccc}
\toprule
\multirow{2}{*}{\textbf{Metric}} & \multicolumn{3}{c|}{\textbf{Homogeneous Scaling}} & \multicolumn{4}{c}{\textbf{Heterogeneous Collaboration}} \\
\cmidrule(lr){2-4} \cmidrule(lr){5-8}
& \small Wheeled$\times$1 & \small Wheeled$\times$3 & \small Wheeled$\times$5 & \small Hum.$\times$1+Wheel.$\times$2 & \small Quad.$\times$1+Wheel.$\times$2 & \small Hum.$\times$1+Quad.$\times$1 & \small Hum.$\times$2+Quad.$\times$2\\
\midrule
AEST (\#) $\downarrow$ & 34.8 & 14.7 \deltag{-58} & 8.5 \deltag{-76} & 16.2 \deltag{-53} & 19.5 \deltag{-44} & 23.0 \deltag{-34} & 10.5 \deltag{-70} \\
SR (\%) $\uparrow$     & 76.6 & 71.7 \deltar{-6}  & 69.7 \deltar{-9} & 72.5 \deltar{-5}  & 71.3 \deltar{-7}  & 73.3 \deltar{-4}  & 70.7 \deltar{-8}  \\
\midrule
\rowcolor[HTML]{DAEFF9} \textbf{SS (\%/\#)} $\uparrow$ & 2.20 & 4.88 \deltag{+122} & 8.20 \deltag{+373} & 4.48 \deltag{+103} & 3.66 \deltag{+66} & 3.19 \deltag{+45} & 6.73 \deltag{+206} \\
\bottomrule
\end{tabular}}
\vspace{-1.5em}
\end{table*}

\textbf{Implementation Details.}
The high-level reasoning in RoboOS-NeXT is driven by the Brain Model, implemented with RoboBrain-2.0~\cite{team2025robobrain}, a multimodal large language model enhanced for spatio-temporal reasoning. It performs global task decomposition, dynamic re-planning, and interaction with STEM. Low-level execution is handled by the Cerebellum Skill Library, which runs on individual robot terminals to translate abstract reasoning into executable actions. In our real-robot demonstrations, this skill library incorporates \emph{navigation} modules based on SLAM techniques and \emph{manipulation} modules based on diffusion-policy~\cite{chi2023diffusion} methods, enabling reliable mobility and contact-rich interaction.

\subsection{Lifelong Adaptability (RQ1)}

To systematically evaluate lifelong adaptability, we categorize tasks across restaurant, supermarket, and household into three levels. 
\textit{Level 1 (Simple)}: directly grounded instructions, local perception, short linear actions, basic skills. 
\textit{Level 2 (Medium)}: local state reasoning, longer sequences with conditionals, coordinated basic or parameterized composite skills. 
\textit{Level 3 (Complex)}: global perception, aggregated reasoning, compound planning with iterative perception–reasoning–action loops. 
In addition to these qualitative distinctions, the levels also differ quantitatively in the number of tree/graph nodes (corresponding to region/carrier nodes in the scene tree, and object nodes in relation graphs): simple tasks typically involve fewer than 20 nodes, medium tasks 20–30 nodes, and complex tasks 40–50 nodes.

We compare RoboOS-NeXT with a \emph{memory-less baseline} that perceives only the current room state, without structured representation or memory updates. Tab.~\ref{tab:lifelong_eval} summarizes results across sequence lengths (SQ) and difficulty levels.
\textbf{\textit{(1) Consistent MSR gains.}} RoboOS-NeXT outperforms the baseline across all domains/levels; under long sequences (SQ=5) the baseline collapses (e.g., Restaurant L2: 0.0\% vs.\ 75.0\%), indicating memory preserves competence over extended horizons. 
\textbf{\textit{(2) Efficiency improves with experience.}} AEST is reduced by 20--70\% versus the baseline; e.g., Household L2 at SQ=5 drops from 41.4 (Baseline) to 15.5 (RoboOS-NeXT, -63\%), showing faster execution as experience accumulates. 
\textbf{\textit{(3) Robust at high complexity.}} Gains persist on L3 tasks (e.g., Supermarket, MSR +63.5\%; Household, +58.3\%) with more than 70\% AEST reductions, demonstrating generalization to global, composite skills.
Overall, RoboOS-NeXT exhibits lifelong adaptability: it maintains stable success while shortening execution across longer sequences and increasing task complexity.

\subsection{Collaborative Scalability (RQ2)}
To assess scalability, we evaluate RoboOS-NeXT across homogeneous and heterogeneous team compositions (Tab.~\ref{tab:scalability_eval}). Three findings emerge.  \textbf{\textit{(1) More agents improve efficiency.}} In homogeneous teams, scaling from 1$\rightarrow$3$\rightarrow$5 wheeled robots reduces AEST by -58\% and -76\% relative to the single-robot baseline, showing near-monotonic efficiency gains from parallelism. \textbf{\textit{(2) Reliability remains stable.}} Despite increased coordination load, SR decreases only modestly in homogeneous scaling (-6\%, -9\%) and in heterogeneous teams (Hum.$\times$1 + Wheel.$\times$2: -5\%). \textbf{\textit{(3) Memory sustains scalability.}} By maintaining shared task context, RoboOS-NeXT converts larger teams into large reductions in execution steps while keeping SR degradation minor, validating that efficiency improvements do not come at the cost of reliability.

\subsection{Scheduling Robustness (RQ3)}
We assess robustness under common error modes spanning three cases: 
E1—\textit{Robot Offline} (disconnection/non-responsiveness), 
E2—\textit{Tool Failure} (loss or malfunction of a capability, e.g., grasping), 
E3—\textit{Brain Model Hallucination} (instructions/decompositions misaligned with the environment).
RoboOS-NeXT is compared to a memory-less baseline that perceives only the current room state without structured representation or memory updates.
As shown in Tab.~\ref{tab:robustness_eval}, three findings emerge.  
\textbf{\textit{(1) Memory is critical.}} RoboOS-NeXT sustains high SR under both \emph{No Error} and all common error modes by re-planning and re-allocating resources.  
\textbf{\textit{(2) The baseline collapses under errors.}} Without memory, SR drops sharply across error types (e.g., E2 to 23.5\%), lacking the context needed for recovery.  
\textbf{\textit{(3) Memory-centric design enables fault tolerance.}} Persisting task context and state yields large gains over the baseline (e.g., E2~+203\%, E3~+153\%), confirming memory as the key enabler of resilient operation.

\begin{table}[!t] 
\centering 
\caption{SR (\%) under common error modes in Household (L1, SQ=1). Performance of RoboOS-NeXT is compared against a memory-less baseline.}
\label{tab:robustness_eval}
\resizebox{0.46\textwidth}{!}
{%
\begin{tabular}{lcccc}
\toprule
\textbf{Settings} & \small No Error & \small E1 & \small E2 & \small E3 \\
\midrule
Baseline & 81.6 & 44.5 & 23.5 & 31.0 \\
RoboOS-NeXT & 89.2 \deltag{+9} & 87.6 \deltag{+97} & 71.3 \deltag{+203} & 78.5 \deltag{+153} \\
\bottomrule
\end{tabular}
}
\end{table}

\begin{table}[!t]
\centering
\caption{Ablation study of STEM components in Household (L1, SQ=1). Results report AEST, SR and SS.}
\label{tab:ablation_study}
\resizebox{0.46\textwidth}{!}{%
\begin{tabular}{lcc|c}
\toprule
\textbf{System Configuration} & AEST(\#)$\downarrow$ & SR(\%)$\uparrow$ & \textbf{SS(\%/\#)$\uparrow$}\\
\midrule
RoboOS-NeXT (Full System) & 11.6 & 89.2 & \textbf{7.69} \\
\midrule
RoboOS-NeXT w/o Spatial Memory & 58.1 & 24.2 & 0.42 \\
RoboOS-NeXT w/o Temporal Memory &  8.7 &  38.3 & 4.40 \\
RoboOS-NeXT w/o Embodiment Memory & -- & 0.0 & -- \\
\bottomrule
\end{tabular}}
\vspace{-1em}
\end{table}

\subsection{Ablation Study (RQ4)}

To examine the contributions of different memory dimensions in STEM, we performed an ablation study by disabling Spatial, Temporal, or Embodiment memory modules in turn and measuring their impact on task execution. As shown in Tab.~\ref{tab:ablation_study}, three conclusions emerge: \textit{\textbf{(1) Spatial memory is essential for efficient exploration.}} Without spatial memory, the system cannot recall previously mapped locations and must repeatedly explore, leading to excessive steps (AEST 58.1) and low success (24.2\%). \textit{\textbf{(2) Temporal memory underpins long-horizon reasoning.}} Without temporal memory, the system loses awareness of prior actions and effectively operates in an open-loop manner; this explains the shorter paths (AEST 8.7) but also the sharp drop in SR (38.3\%). \textit{\textbf{(3) Embodiment memory is indispensable for multi-robot coordination.}} Without embodiment-level awareness, the system cannot ground actions to specific robots or synchronize their roles, resulting in complete task failure (SR 0.0). These confirm that the synergy of spatial, temporal, and embodiment memory is crucial for RoboOS-NeXT’s overall capability.

\begin{figure}[t]
    \centering
    \includegraphics[width=0.92\linewidth]{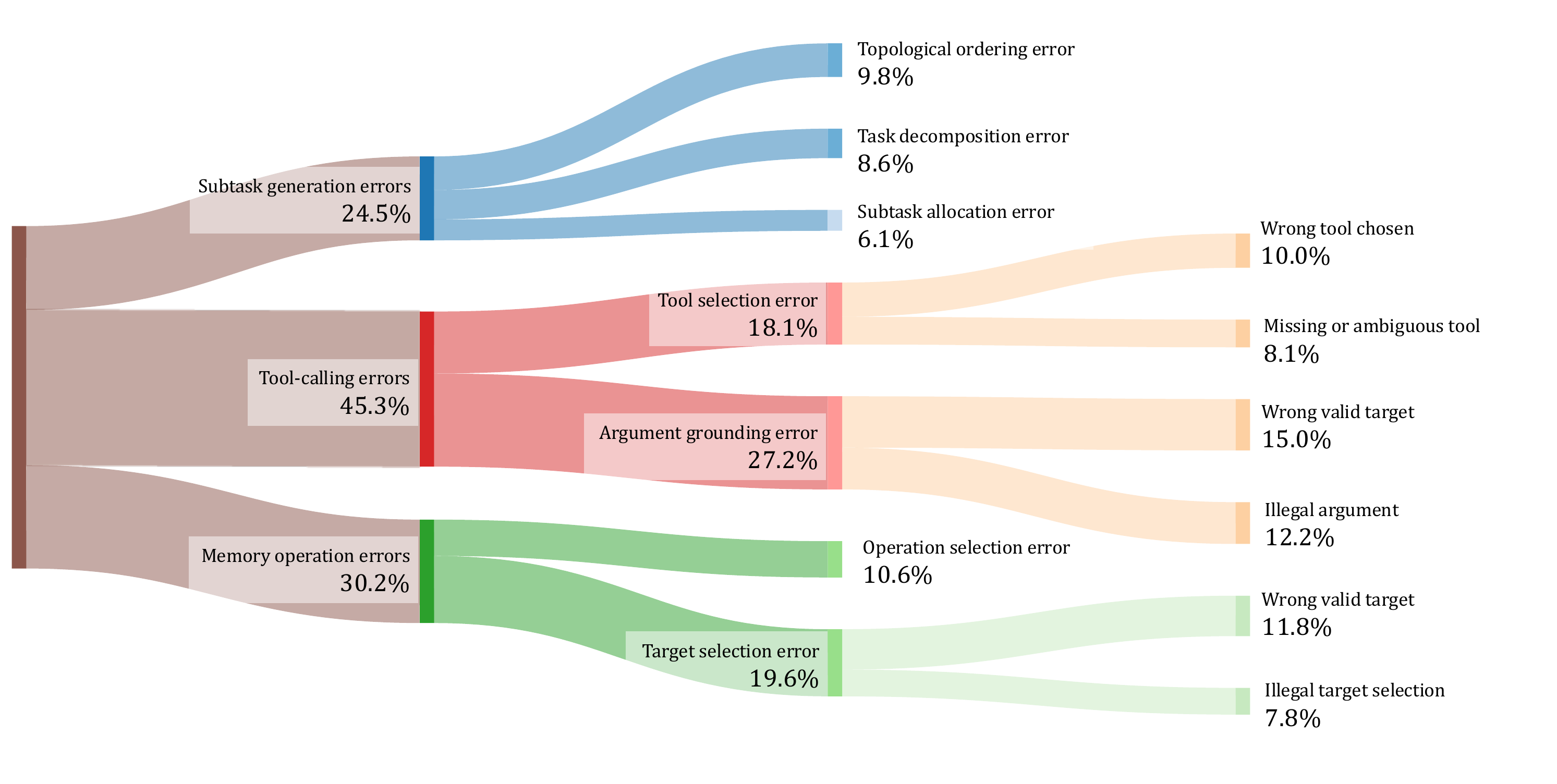}
    \caption{\textbf{Failure distribution in the restaurant scenario.} Most errors arise from tool invocation and memory operations, with additional sensitivity in subtask generation.}
    \label{fig:failure_distribution}
    \vspace{-1em}
\end{figure}

\subsection{Failure Analysis (RQ5)}
We analyzed 53 failures across 200 trials in the restaurant scenario (Fig.~\ref{fig:failure_distribution}) and identified three dominant sources as follow. \textbf{\textit{(1) Subtask generation error (24.5\%).}} Complex or ambiguous task graphs induce misordered dependencies and coarse decompositions, revealing sensitivity to structural priors. \textbf{\textit{(2) Tool invocation error (45.3\%).}} Errors are dominated by brittle parameter binding (e.g., navigation/grasp targets drifting to nearby objects), indicating insufficient semantic alignment between memory, perception, and control. \textbf{\textit{(3) Memory operation error (30.2\%).}} Over long horizons, noise in update/selection accumulates, degrading temporal consistency.
Overall, failures cluster around structured reasoning and long-horizon consistency rather than missing primitives. Strengthening task-graph regularization, improving grounding for parameterized tools, and refining memory update/retrieval mechanisms are promising directions for enhancing RoboOS-NeXT robustness.

\subsection{Demonstrations in Real-World Collaboration}  
\begin{figure}[!htb]
    \centering
    \includegraphics[width=0.98\linewidth]{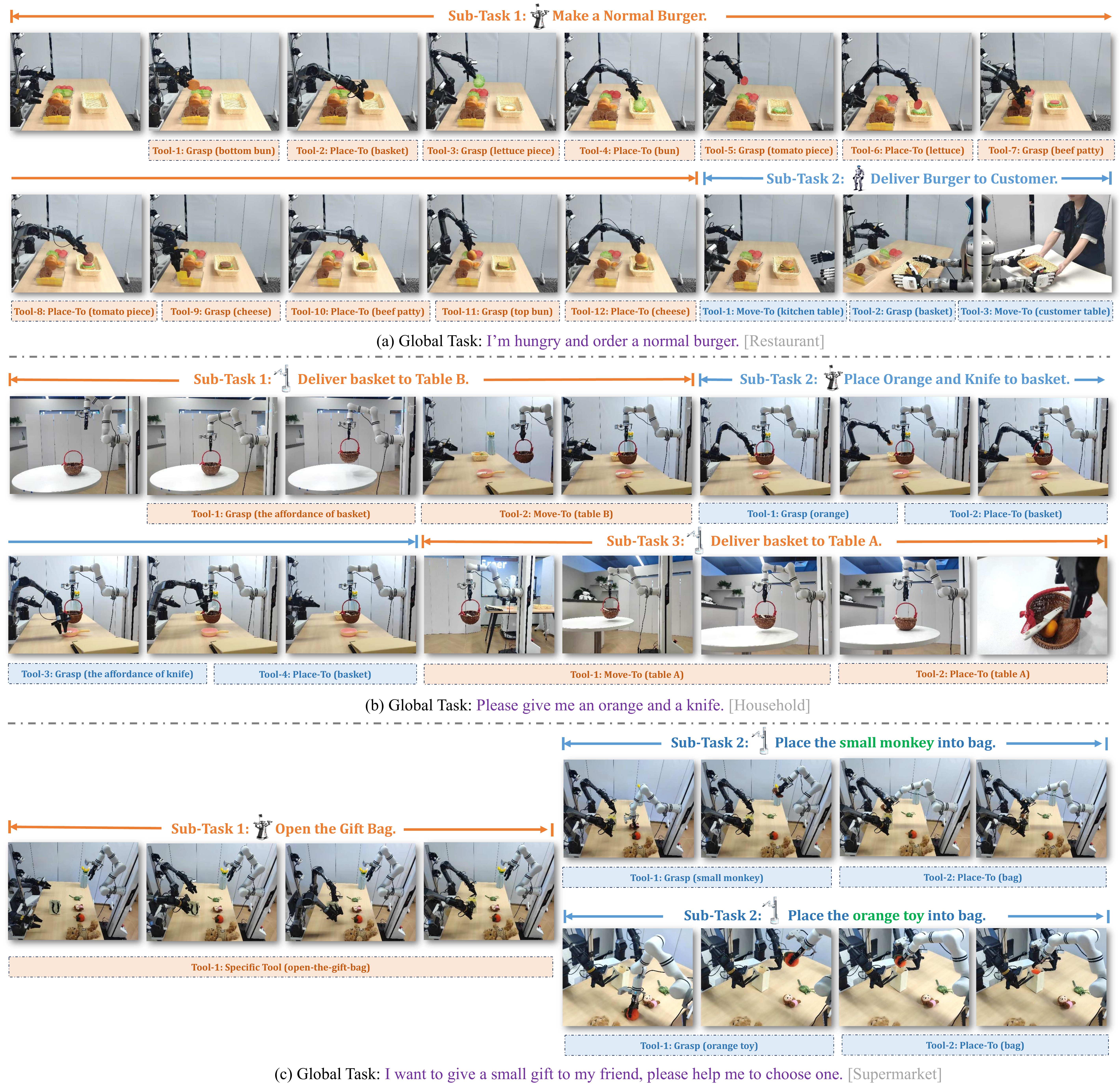}
    \caption{
        \textbf{Real-world RoboOS-NeXT Demonstrations.} 
        We showcase multi-robot collaboration in three types of scenarios: 
        (a) Restaurant, 
        (b) Household and 
        (c) Supermarket.
    }
    \label{fig:demo}
    \vspace{-1.5em}
\end{figure}

We validate RoboOS-NeXT in three real-world collaboration scenarios: restaurant, household, and supermarket. In the restaurant setting (Fig.~\ref{fig:demo}~(a)), a Unitree G1 humanoid and an Agilex dual-arm robot respond to the request, ``I'm hungry and order a normal burger.'' The robotic brain model decomposes this instruction into subtasks for burger preparation and delivery, assigning roles to each robot. In the household setting (Fig.~\ref{fig:demo}~(b)), a Realman single-arm and an Agilex dual-arm robot jointly fetch items such as ``an orange and a knife,'' handling both parallel and sequential dependencies. In the supermarket (Fig.~\ref{fig:demo}~(c)), RoboOS-NeXT supports gift selection and packaging: the brain model reasons about dimensions and bag compatibility, the Agilex opens the bag, and the Realman places the gift inside. These demonstrations highlight RoboOS-NeXT’s ability to bridge high-level reasoning and low-level execution in heterogeneous teams, and point toward extensions to more complex multi-robot collaborations.

\section{Conclusions}
In this paper, we introduced \textbf{RoboOS-NeXT}, a memory-based framework for multi-robot collaboration. At its core is the Spatio-Temporal–Embodiment Memory (STEM), which unifies spatial, temporal, and embodiment information into a shared representation. Coupled with a brain–cerebellum framework, RoboOS-NeXT forms a closed loop between reasoning and execution, enabling synchronized coordination and fault-tolerant operation.  
Our evaluation across diverse tasks and embodiments demonstrates that RoboOS-NeXT provides a principled foundation for \textit{lifelong adaptability}, \textit{scalable collaboration}, and \textit{robust scheduling}, marking a step toward more general and reliable embodied intelligence.

\bibliographystyle{IEEEtran}
\bibliography{main}


\addtolength{\textheight}{-12cm}   

\end{document}